\documentclass[10pt,twocolumn,letterpaper]{article}

\usepackage{cvpr}
\usepackage{times}
\usepackage{epsfig}
\usepackage{graphicx}
\usepackage{amsmath}
\usepackage{amssymb}
\usepackage{algorithm}
\usepackage{algpseudocode}
\usepackage{subcaption}
\usepackage{booktabs}
\usepackage{multirow}

\usepackage[pagebackref=true,breaklinks=true,letterpaper=true,colorlinks,bookmarks=false]{hyperref}

\cvprfinalcopy 

\def\cvprPaperID{7022} 

\ifcvprfinal\fi

\begin{document}

\title{Better Captioning with Sequence-Level Exploration}

\author{Jia Chen\\
Carnegie Mellon University\\
Language Technology Institute\\
{\tt\small marshimarocj@gmail.com}
\and
Qin Jin\thanks{corresponding author}\\
Renmin University of China\\
School of Information\\
{\tt\small qjin@ruc.edu.cn}
}

\maketitle

\begin{abstract}
Sequence-level learning objective has been widely used in captioning tasks to achieve the state-of-the-art performance for many models. 
In this objective, the model is trained by the reward on the quality of its generated captions (sequence-level). 
In this work, we show the limitation of the current sequence-level learning objective for captioning tasks from both theory and empirical result. 
In theory, we show that the current objective is equivalent to only optimizing the precision side of the caption set generated by the model and therefore overlooks the recall side. 
Empirical result shows that the model trained by this objective tends to get lower score on the recall side. 
We propose to add a sequence-level exploration term to the current objective to boost recall. 
It guides the model to explore more plausible captions in the training. 
In this way, the proposed objective takes both the precision and recall sides of generated captions into account. 
Experiments show the effectiveness of the proposed method on both video and image captioning datasets. 
\end{abstract}

\section{Introduction}
\label{sec:introduction}
Captioning is one of the core tasks in vision and language fields. 
The input is an image or video and the output is a descriptive sentence. 
In terms of the output structure, the descriptive sentence is actually a sequence, which is more complex than the output of classification and detection tasks and therefore poses a challenge for the learning objective in captioning tasks. 
Furthermore, there exists multiple correct captions for the same input and it is impossible to enumerate all the correct captions when collecting the groundtruth. 
The above two unique properties, sequence structure and multiple correct grountruth captions, make captioning tasks difficult and worth special treatment for its own learning/training objective. 

\begin{figure}
    \centering
    \includegraphics[width=.9\linewidth]{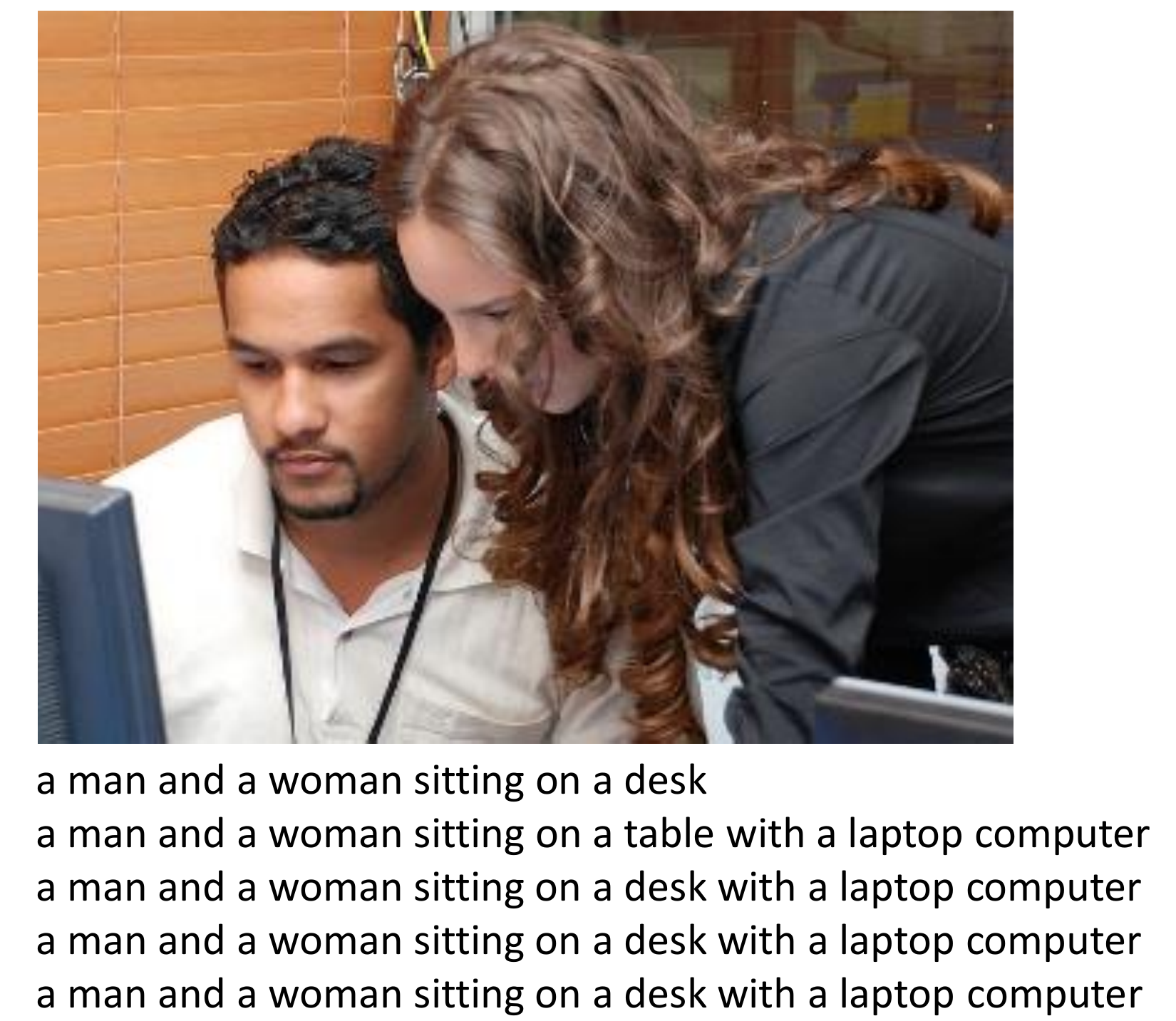}
    \caption{Illustration on limitations of current sequence-level learning: $5$ captions randomly sampled from the model \cite{self_critique} are almost identical, which indicates that the model is not likely to have high recall. }
    \label{fig:introduction}
\end{figure}

Most caption models \cite{google, self_critique,bottom_up} are based on the encoder-decoder architecture and we will only talk about training objectives associated with this architecture. 
The original training objective is cross-entropy loss \cite{google}, which does word-level supervision. 
To be specific, the decoder is fed with the word from the groundtruth caption at each step and predicts the word at next step. 
Thus, the decoder is trained to focus on the correctness of predicting each word separately. 
However, at each step in the test stage, the decoder is fed with the word predicted from the previous step rather than the groundtruth word. 
This leads to the gap between training and test and limits the performance in the test. 
Later, sequence-level learning objective is proposed by researchers to address this gap \cite{mixer, self_critique}. In this objective, only after the whole sentence is generated by the decoder, the quality of the caption is evaluated by a score and that score is used to guide the model training. 
That is, the decoder predicts the word at each step based on the word predicted at last step in both training and test stages. 
The sequence-level learning objective \cite{mixer, self_critique} is shown to improve performance significantly on most evaluation metrics such as CIDEr\cite{cider}, METEOR\cite{meteor} and SPICE\cite{spice} compared to the cross-entropy loss. 

In this paper, we show the limitations of the current sequence-level learning objective from both theoretical and empirical aspects despite its success in captioning tasks. 
From theoretical aspect, we show that the current objective is equivalent to optimizing the precision side of the predicted caption set. 
The standard precision is defined based on the set membership of an element. 
And the set membership function outputs 0-1 for a caption, which describes whether the caption belongs to a set or not. 
We relax the 0-1 set membership function used in precision calculation to real-value output within range $[0, 1]$. 
The relaxed set membership function describes the confidence of a caption belonging to a set. 
In this way, we show that the current sequence-level learning objective is equivalent to maximizing the generalized precision with the relaxed set membership function and it overlooks the recall side of the problem. 
From empirical aspect, we show that the model trained by the current sequence-level learning objective tends to cover very few different captions in its predictions and gets low score on recall related metrics. 
As illustrated in figure~\ref{fig:introduction}, we randomly sample $5$ sentences from the model and the resulting $5$ sentences are almost identical. 

To overcome the limitations of the current sequence-level learning objective, we propose to add a sequence-level exploration term to boost recall. 
In this exploration term, we maximize the difference between the generated captions (sequence-level) of the same input.
One example of difference measurement could be edit distance. 
In the context of captioning task, the proposed exploration term corresponds to maximizing the diversity \cite{caption_gan} of generated captions. 
Furthermore, we show that diversity is a proxy measurement of recall for captioning. 
In training, this term encourages the model to explore more different captions. 
Such sequence-level exploration is different from the typical maximum-entropy exploration regularization \cite{entropy_rl} that is put on the policy in reinforcement learning. 
In typical maximum-entropy exploration regularization, it maximizes the uncertainty of the policy at each step. 
That is, given generated words up to step $t$, it maximizes the uncertainty of the next word. 
We call this word-level exploration. 

In summary, the contributions of this work are:\\*
1) We show the limitations of the current sequence-level learning objective for the captioning task from both theoretical and empirical aspects. \\*
2) We propose a new learning objective for the captioning task which adds a sequence-level exploration term to boost recall. \\*
3) The derived solution from the proposed objective achieves better performance on various standard evaluation metrics of the precision side. It also improves the performance on recall related metrics. 


\section{Related Work}
\label{sec:related}
The dominant neural network architecture of the captioning task is based on the encoder-decoder framework \cite{bahdanau2014neural}. 
Early works \cite{google, baidu, s2s} use convolution neural network as encoder and recurrent neural network with LSTM cell \cite{lstm} as decoder. 
In the image captioning task, Xu et al. \cite{spatial_attention} proposed the spatial attention, which selects relevant image regions to generate image descriptions. 
In the video captioning task, Yao et al. \cite{temporal_attention} proposed the temporal attention, which expands the attention mechanism in the temporal direction. 
After that, different variants of attention mechanism are proposed to further improve the performance, such as attention on semantic concepts \cite{semantic_attention,pan2017video,cvpr2} and adaptive attention on visual and linguistic contexts \cite{adjusted_attention,adaptive_attention,cvpr1}. 
The latest variation on attention mechanism is the up-down attention \cite{bottom_up} which enables attention to be calculated at the level of objects and other salient image regions. 
In addition to attention mechanism, researchers also propose other modification on the neural network architecture. 
Pan et al. \cite{pan2015hierarchical} utilized the hierarchical encoder to learn better visual representations. 

The original objective function \cite{google, baidu} used in the captioning task is cross-entropy loss, which applies word-level supervision. 
To be specific, in training, the model is fed with the groundtruth word at each step and supervision monitors whether the model outputs the correct next word. 
We call such supervision as word-level supervision. 
However, in the test stage, the model is fed with the word predicted by itself at last step rather than the groundtruth word. 
This is known as the train-test gap in sequence prediction tasks. 
Bengio et al. \cite{scheduled_sampling} proposed scheduled sampling, a curriculum learning approach, to minimize such gap. 
Later, sequence-level training is proposed by Ranzato et al. \cite{mixer} to systematically address this issue. 
Different from word-level supervision, the sequence-level learning evaluates the sentence only after the whole sentence has been generated. 
The sentence is evaluated by a reward about its semantic coherence with the groundtruth caption.
And the reward is usually set to be the evaluation metric that has high correlation with human judgement. 
Rennie et al. \cite{self_critique} further improves the sequence-level learning by introducing a special baseline in reward, which is the score of the caption greedily decoded from the current model. 
Sequence-level training objective has been widely used in captioning tasks to achieve state-of-the-art performance \cite{bottom_up,discriminative,qjin1,qjin2}. 

\section{Limitations of Current Sequence-level Learning}
\label{sec:limitation}
In this section, we show the limitation of current sequence-level learning for the captioning task from both theoretical and empirical aspects. 
Theoretically, we show that the current objective function of sequence-level training is equivalent to optimizing the generalized precision with relaxed set membership function on the predicted captions. 
Empirically, we show that the model trained by the current sequence-level learning tends to generate very few different captions for the same input and does not get high score on recall related metrics. 

\subsection{Limitation from theory}
We first relax the set membership function in the standard precision measurement for the captioning task. 
Then we show that the objective of current sequence-level learning is actually optimizing the generalized precision with relaxed set membership function in the context of captioning task. 

Suppose that the space of all the possible sentences is $\mathcal{Y}$, the groundtruth sentence set of an input (image / video) $x_i$ is $Y$ and the predicted sentence set of that input by the captioning model is $\widetilde{Y}$. 
Then the precision is defined by:
\begin{align}
Precision(Y, \widetilde{Y}) &= \frac{|Y \cap \widetilde{Y}|}{|\widetilde{Y}|}\nonumber\\
 &= \frac{\sum_{y \in \mathcal{Y}}\delta[y \in Y] \delta[y \in \widetilde{Y}]}{\sum_{y \in \mathcal{Y}}\delta[y \in \widetilde{Y}]}\nonumber\\
 &= \sum_{y \in \mathcal{Y}} \delta[y \in Y] \underbrace{\frac{\delta[y \in \widetilde{Y}]}{\sum_{y' \in \mathcal{Y}}\delta[y' \in \widetilde{Y}]}}_{p(y \in \widetilde{Y})}\nonumber\\
 &= \sum_{y \in \mathcal{Y}} \delta[y \in Y] p(y \in \widetilde{Y})\label{eq:precision}
\end{align}
Inside the summation of eq~\eqref{eq:precision}, it contains two terms: $\delta[y\in Y]$ and $p(y \in \widetilde{Y}) =\frac{\delta[y \in \widetilde{Y}]}{\sum_{y' \in \mathcal{Y}}\delta[y' \in \widetilde{Y}]}$. 
In the $\delta[y \in Y]$ term, the $\delta$ function checks whether or not caption $y$ belongs to groundtruth sentence set $Y$. 
In the $p(y \in \widetilde{Y})$ term, the $\delta$ function checks whether or not caption $y$ belongs to the predicted sentence set $\widetilde{Y}$. 

For the $\delta[y \in Y]$ term,  we relax the binary valued $\delta$ function to a real-valued function $\Delta(y, Y)$ with output in the range of $[0, 1]$:
\begin{equation}
\label{eq:first}
\delta[y \in Y] \to \Delta(y, Y)
\end{equation}
$\Delta(y, Y)$ indicates the likelihood of each individual $y$ within the set $Y$ and is a relaxed set membership function. 
One natural choice for  $\Delta(y, Y)$ is to use the evaluation metric normalized by its maximum value. 
As all the current evaluation metrics in the captioning task are bounded, they can be normalized properly. 
For simplicity, we assume that we are dealing with the evaluation metric $\Delta(y, Y)$ that has already been normalized. 

The term $p(y \in \widetilde{Y})$ can be interpreted as the chance of the sentence $y$ within set $\widetilde{Y}$. 
Note that the value of $\delta[y \in \widetilde{Y}]$ is 0-1, which represents whether the captioning model considers sentence $y$ as correct or not. 
Correspondingly, $p(y \in \widetilde{Y})$ can only take values eithor $0$ if $y \notin \widetilde{Y}$ or $\frac{1}{|\widetilde{Y}|}$ if $y \in \widetilde{Y}$. 
It does not cover the whole range $[0, 1]$ of a probability. 
If we again relax the 0-1 membership function $\delta[y \in \widetilde{Y}]$ to a real-valued confidence, $p(y \in \widetilde{Y})$ can cover the whole range $[0, 1]$ of a probability. 
After the relaxation, $p(y \in \widetilde{Y}) $ is actually the probability of caption $y$ from the captioning model. 
Thus by using the relaxed set membership function, we replace $p(y \in \widetilde{Y})=\frac{\delta[y \in \widetilde{Y}]}{\sum_{y' \in \mathcal{Y}}\delta[y' \in \widetilde{Y}]}$ with $p_\theta(y|x_i)$, which is the probability from the captioning model:
\begin{equation}
\label{eq:second}
p(y \in \widetilde{Y}) = \frac{\delta[y \in \widetilde{Y}]}{\sum_{y' \in \mathcal{Y}} \delta[y' \in \tilde{Y}]} \to p_\theta(y|x_i)
\end{equation}

Substituting $\delta[y \in Y]$ and $p(y \in \widetilde{Y})$ in eq~\eqref{eq:precision} by \eqref{eq:first} and \eqref{eq:second} respectively, we get the generalized precision (GP) for the captioning task:
\begin{equation}
\label{eq:gp}
GP(Y, \theta|x_i) = \sum_{y \in \mathcal{Y}} \Delta(y, Y) p_\theta(y|x_i)
\end{equation}
We could use generalized precision $GP$ to rewrite the original sequence-level learning objective for the captioning task. 
Setting $\Delta(y, Y)$ as reward, the original objective is to maximize the expected return:
\begin{equation}
\label{eq:return}
J(\theta) = \sum_{i = 1}^n \mathbb{E}_{p_\theta(y|x_i)}\Delta(y, Y) 
\end{equation}
By comparing eq~\eqref{eq:return} with the generalized precision measurement defined in eq~\eqref{eq:gp}, we see that they are exactly the same: 
\begin{align}
\label{eq:equal}
\begin{split}
J(\theta) &= \sum_{i=1}^n \sum_{y \in \mathcal{Y}} \Delta(y, Y) p_\theta(y|x_i)\\
&= \sum_{i=1}^n GP(Y, \theta|x_i)
\end{split}
\end{align}
This means that sequence-level learning objective only optimizes the precision side of the captions predicted by the captioning model. 
However, as there exist multiple correct answers for the same input $x_i$, which means that the recall side should also be taken into account when training the captioning model. 
On the contrary, the original objective totally overlooks the recall side of the problem. 

\subsection{Limitation from empirical results}
Complementary to the theoretical analysis above, we also measure the precision and recall side of the model trained by current sequence-level learning objective. 
The precision side could be measured by the standard evaluation metrics in captioning tasks such as METEOR\cite{meteor} and SPICE\cite{spice}. 
As it is not possible to collect all the correct answers for an input $x_i$, directly computing recall is not feasible. 
Instead, we use set level diversity metrics \cite{caption_gan} \emph{Div-1}, \emph{Div-2} and \emph{mBleu} as a proxy measurement of the recall. 
The set level diversity metrics are defined on a set of captions, $\widetilde{Y}$, corresponding to the same input $x_i$. 
\begin{itemize}
    \item \emph{Div-1} ratio of the number of unique unigrams in $\widetilde{Y}$ to the number of words in $\widetilde{Y}$. Higher is more diverse. 
    \item \emph{Div-2} ratio of the number of unique bigrams in $\widetilde{Y}$ to the number of words in $\widetilde{Y}$. Higher is more diverse. 
    \item \emph{mBleu} Bleu score is computed between each caption in $\widetilde{Y}$ against the rest. Mean of these Bleu scores is the mBleu score. Lower is more diverse. 
\end{itemize}
To report set level diversity metrics, we sample $5$ captions from the model for each input. 
Correspondingly, when calculating the precision metric CIDEr, we average the CIDEr scores of the $5$ sampled captions.

Here is the reasoning of why the above diversity metrics is related to recall. 
Standard recall is defined by:
\begin{align}
\begin{split}
Recall(Y, \widetilde{Y}) &= \frac{|Y \cap \widetilde{Y}|}{Y}\\
& \propto |Y \cap \widetilde{Y}|\\
& \propto  |\widetilde{Y}| Precision(Y, \widetilde{Y})
\end{split}
\end{align}
When the precision is fixed, we see that the recall is proportional to the size of the predicted set $\widetilde{Y}$. 
To compare the recall at the same precision level, we could instead compare the size of the predicted caption set from the model. 
In this way, any measurement on the size of set $\widetilde{Y}$ could be considered as a proxy measurement of recall. 
Directly measuring the size of $\widetilde{Y}$ by the number of captions is not meaningful if we are allowed to sample infinite times from the model. 
A more meaningful way to measure the size of $\widetilde{Y}$ is:
\emph{given fixed number of sampling times, calculating the difference between sampled captions}. 
And this is exactly the quantity defined in set level diversity metrics. 

\begin{table}[!t]
\centering
\caption{Comparison between word-level cross-entropy loss (XE) and sequence-level learning (SLL) on precision and recall sides}
\label{tab:empirical}
\begin{tabular}{ccccc}\toprule
\multirow{2}{*}{Method} & Precision & \multicolumn{3}{c}{Recall} \\
& CIDEr ($\uparrow$) & Div1 ($\uparrow$) & Div2 ($\uparrow$) & mBleu4 ($\downarrow$)  \\\cmidrule(lr){1-1}\cmidrule(lr){2-2}\cmidrule(lr){3-5}
XE & 74.2 & $0.57$ & $0.78$ & $0.06$ \\
SLL & 114.6 & $0.25$ & $0.32$ & $0.81$ \\\bottomrule
\end{tabular}
\end{table}

\begin{figure}
    \centering
    \includegraphics[width=\linewidth]{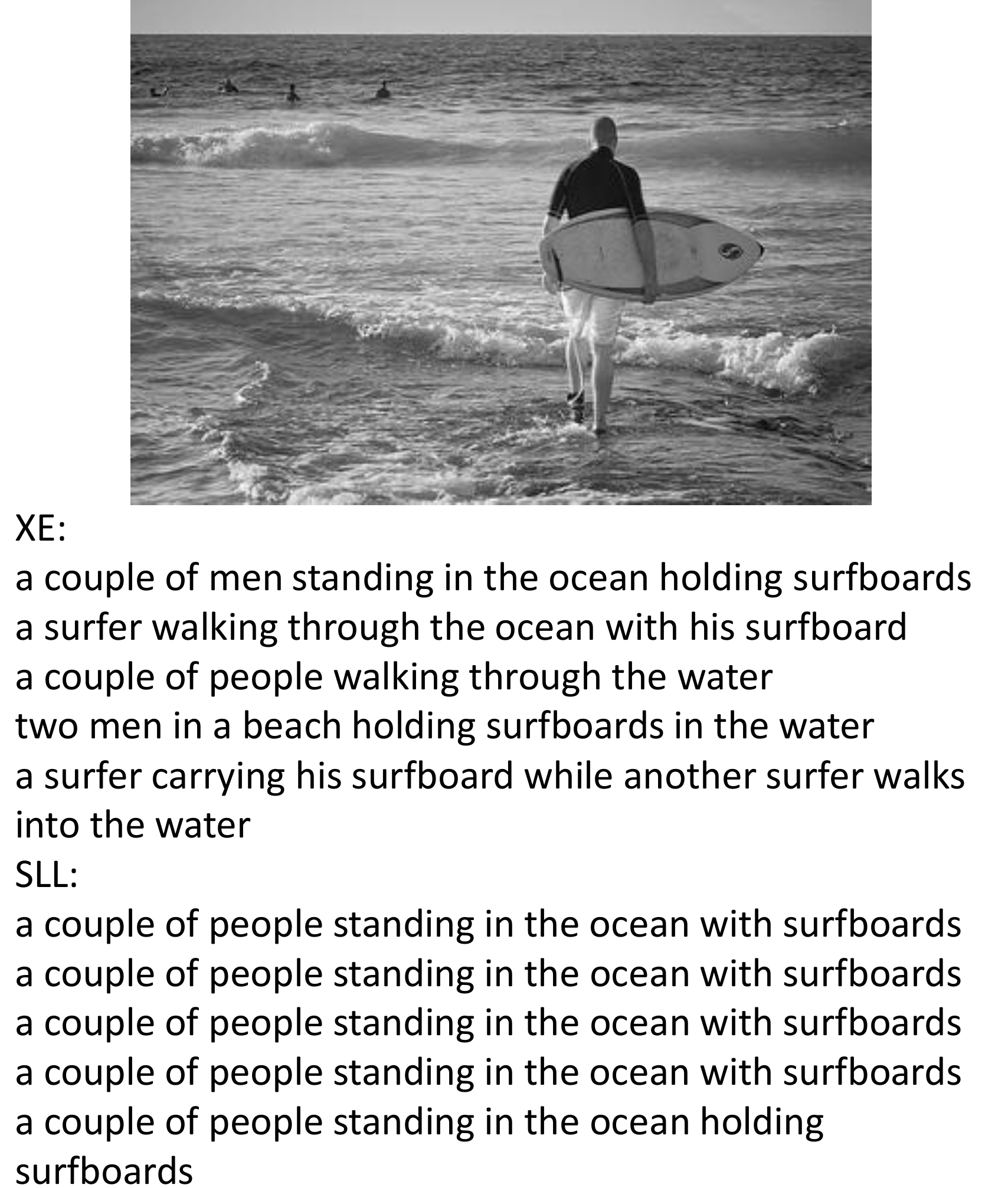}
    \caption{Illustration of $5$ captions sampled from models given the same input: XE is the model trained by cross-entropy objective and SLL is the model trained by sequence-level learning objective. }
    \label{fig:empirical}
\end{figure}

\begin{figure}
\centering
\includegraphics[width=0.9\linewidth]{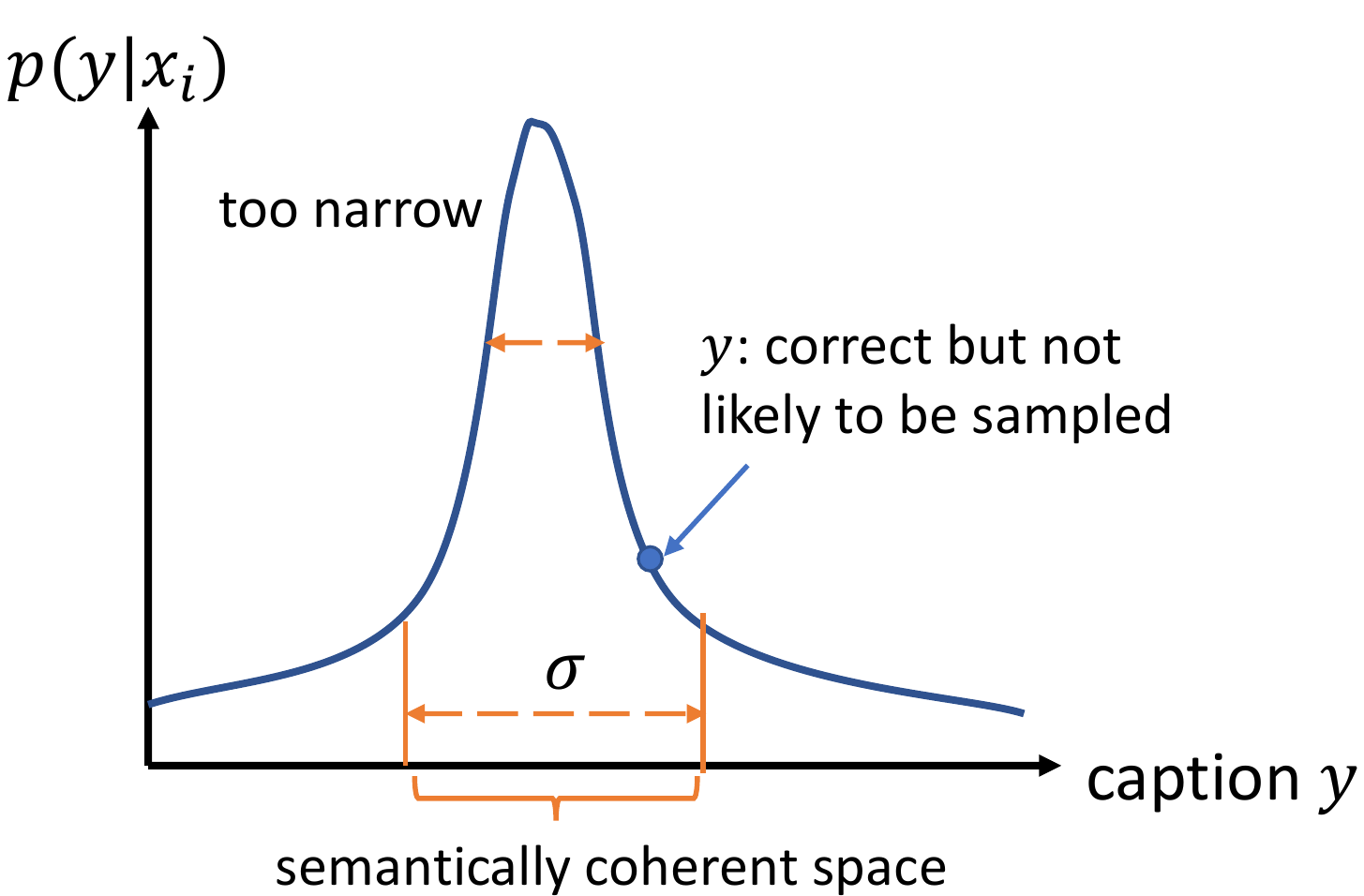}
\caption{Illustration of the peak width of caption distribution $p(y|x)$ based on empirical results of the sequence-level learning objective}
\label{fig:illustrate_p}
\end{figure}

As shown in table~\ref{tab:empirical} compared to word-level cross-entropy (XE) loss, sequence-level learning (SLL) leads to a large performance drop on the recall side though it improves the metrics on the precision side significantly. 
This could be further illustrated by the examples shown in figure~\ref{fig:empirical}. 
In this example, $5$ randomly sampled captions are almost identical for the model trained by sequence-level learning (SLL) objective while this is not an issue for the model trained by the word-level cross-entropy (XE) objective. 
We explain this observation by the peak width of the distribution. 
As illustrated in figure~\ref{fig:illustrate_p}, suppose we project the captions to a one-dimensional space and the width of the line segment containing semantic coherent captions for an input $x_i$ is $\sigma$. 
Based on the empirical result observed in this section, the peak width of the model trained by SLL objective should be much smaller than $\sigma$ so that most sampled sentences for input $x_i$ are almost identical. 
However, the peak width of an ideal model should be similar to $\sigma$. 
In this case, the samples from the model is likely to cover the semantically coherent space and get high score on recall as a result. 

\section{Solution}
\label{sec:solution}
We first propose a new objective function to address the limitations of current sequence-level learning objective shown in the last section. 
Then we derive the optimization procedure for this new objective function. 
Finally, we describe the network architecture and training details in implementation. 

\subsection{Objective Function}
As we have shown that diversity is a proxy measurement of recall, we introduce an additional diversity term to the original sequence-level learning objective function to cover the recall side of the problem: 
\begin{align}
\label{eq:problem}
\begin{split}
\text{max}_\theta: &\alpha \underbrace{\sum_{y \in \mathcal{Y}} \Delta(y, y_i) p_\theta(y|x_i)}_{\text{precision}} + \\
& (1-\alpha) \underbrace{\sum_{y \in \mathcal{Y}}\sum_{y' \in \mathcal{Y}} d(y, y') p_\theta(y|x_i) p_\theta(y'|x_i)}_{\text{diversity}}
\end{split}
\end{align}
In this objective function, $x_i$ is the input image or video, $y_i$ is the groundtruth caption, $y$ and $y'$ are any two captions in the caption space $\mathcal{Y}$ that can be sampled from the caption model. 
$p_\theta(y|x_i)$ is the conditional probability given by the caption model. \\*
$\bullet$  $\Delta(y, y_i)$ in precision term measures semantic coherence between caption $y$ and the groundtruth caption $y_i$. 
It is equivalent to $\Delta(y, Y)$ when there is only one groundtruth caption $y_i$ of input $x_i$. 
It encourages the model to put more probability mass $p_\theta(y|x_i)$ on captions that is semantically coherent with the groundtruth. 
Example choices for $\Delta(y, y_i)$ could be METEOR, CIDEr, SPICE, which are shown to have good correlation with human judgements. \\*
$\bullet$ $d(y, y')$ in diversity term measures the syntactic difference between two captions. 
It encourages the model to explore more different ways to express the same semantic meaning. 
Example choices for $d(y, y')$ could be edit distance or BLEU3/4, which measures the difference in sentence structure.

The diversity term is different from the standard maximum-entropy regularization used in reinforcement learning \cite{entropy_rl}, which is put on the \emph{policy} by $\mathbb{H}(p_\theta(w_j|w_{<j}, x_i))$ and maximizes the uncertainty of the next step word $w_j$ given the past words $w_{<j}$. 
The diversity term introduced here is directly put on captions, which are \emph{trajectories} in the reinforcement learning. 
Furthermore, we use distance $d$ rather than entropy of captions to avoid the intractable estimation of denominator $Z$ that involves summing over the probability of all captions. 
Using distance $d$ also offers us more flexibility to plug-in any measurement of difference in sentence structure. 
Thus, compared to standard maximum-entropy regularization, the diversity term has more direct effect on encouraging the model to explore more different captions and is more flexible for more syntactic difference measurements. 

Putting both precision term and diversity term together, the meaning of the proposed objective function is to encourage the model to 
\emph{explore more captions different in syntax but are semantically coherent with the groundtruth caption $y_i$ of input $x_i$. }
Hyper-parameter $\alpha$ is introduced to balance between precision and diversity terms. 

\subsection{Optimization}
We first show that the precision term in the objective function could be directly solved using REINFORCE algorithm \cite{rl}.
Then we show that the diversity term could be solved with some variation on the technique used in the REINFORCE algorithm. 
Finally, we derive the surrogate loss and a complete algorithm for our objective function. 

In optimization convention, we always minimize the objective function. 
Thus, we take negation of the objective function in eq~\eqref{eq:problem} and decompose it into two parts:
\begin{align}
\begin{split}
L(\theta) &= \alpha L_1(\theta) + (1-\alpha) L_2 (\theta)\\
L_1(\theta) &= -\sum_{y \in \mathcal{Y}} \Delta(y, y_i) p_\theta(y|x_i)\\
L_2(\theta) &= -\sum_{y \in \mathcal{Y}} \sum_{y' \in \mathcal{Y}} d(y, y') p_\theta(y|x_i) p_\theta(y'|x_i)
\end{split}
\end{align}
\emph{1. Solution to $L_1(\theta)$: }
We could rewrite $L_1$ as expectation:
\begin{align}
\begin{split}
L_1(\theta)&=-\sum_{y \in \mathcal{Y}} \Delta(y, y_i) p_\theta(y|x_i)\\
&=-\mathbb{E}_{p_\theta (y|x_i)} [\Delta(y, y_i)]
\end{split}
\end{align}
We could use REINFORCE \cite{rl} to calculate its gradient: 
\begin{align}
\begin{split}
\nabla L_1(\theta) &= -\mathbb{E}_{p_\theta (y|x_i)}[\Delta(y, y_i)\nabla \log p_\theta(y|x_i)]\\
&\approx -\Delta(\tilde{y}, y_i) \nabla \log p_\theta(\tilde{y}|x_i)
\end{split}
\end{align}
The second line is Monte Carlo sampling with just one sample caption $\widetilde{y}$ from the model. \\*
\emph{2. Solution to $L_2(\theta)$: }
we could also rewrite $L_2$ as expectation:
\begin{align}
\begin{split}
L_2(\theta) &= -\sum_{y \in \mathcal{Y}}\sum_{y' \in \mathcal{Y}} d(y, y') p_\theta(y|x_i) p_\theta(y'|x_i)\\
=& -\mathbb{E}_{p_\theta(y|x_i)}\mathbb{E}_{p_\theta(y'|x_i)} d(y, y') 
\end{split}
\end{align}
We see that there are two expectations involved. 
We could still apply REINFORCE to the outer expectation and inner expectation respectively and get:
\begin{align}
\begin{split}
\nabla L_2(\theta) &= -\mathbb{E}_{p_\theta (y'|x_i)}\Big[ \mathbb{E}_ {p_\theta(y|x_i) } [d(y, y')] \nabla \log p_\theta(y|x_i)\Big] \\
& -\mathbb{E}_{p_\theta (y|x_i)}\Big[ \mathbb{E}_{p_\theta (y'|x_i)} \big[d(y, y')\nabla \log p_\theta(y'|x_i)\big]\Big]
\end{split}
\end{align}
Approximating it by Monte Carlo sampling leads to the following solution: 
we sample $s$ captions $\widetilde{y}_1, \dots, \widetilde{y}_s$ and calculate pairwise distances. 
For each sample $\widetilde{y}_j$, its corresponding gradient is:
\begin{equation}
\nabla L_2 (\theta) = -\frac{2}{s^2} \sum_{j=1}^s \big( \sum_{k=1}^s d(\widetilde{y}_j, \widetilde{y}_k) \nabla \log p_\theta(\widetilde{y}_j|x_i) \big)
\end{equation}
\emph{3. Complete solution: }
In standard policy gradient of reinforcement learning, the multiplier before $\nabla \log p_\theta (\widetilde{y}_j|x_i)$ represents the reward. 
In the gradient of $L_2$, the multiplier is $\sum_{k=1}^s d(\widetilde{y}_j, \widetilde{y}_k)$ for each sample $\widetilde{y}_j$. 
It is the sum of sample $\widetilde{y}_j$'s distance to other samples of input $x_i$. 
This aligns exactly with our formulation of $L_2$, which is the diversity term. 
This multiplier could be further considered as ``reward'' that involves multiple samples of the input $x_i$ jointly in calculation while calculating standard reward only uses each sample separately. 

Finally, we wrap up all the gradients of $L(\theta)$ in the following surrogate loss of the entire stochastic computation graph \cite{stochastic_computation_graph}: 
\begin{align}
    \mathcal{L}(\theta) = &\frac{1}{s}\sum_{j=1}^s\mathcal{L}^j(\theta) \label{eq:surrogate_loss}\\
    \mathcal{L}^j(\theta) = & -\alpha \Delta(\widetilde{y}_j, y_i) \log p_\theta(\widetilde{y}_j|x_i) \label{eq:surrogate_loss_sample}\\
    &-(1-\alpha) \frac{2}{s} \sum_{k=1}^s d(\widetilde{y}_j, \widetilde{y}_k) \log p_\theta(\widetilde{y}_j|x_i) \nonumber
\end{align}
Following the standard procedure in sequence-level learning of the captioning task, we first train the model by the word-level cross-entropy loss and then switch to this surrogate loss for training. 
Algorithm~\ref{alg:alg} summarizes the entire training process. 

\begin{algorithm}
\caption{Training algorithm of sequence-level exploration}
\label{alg:alg}
\begin{algorithmic}[1]
\For{epoch in [0, M)}
    \State train by cross-entropy loss
\EndFor
\For{epoch in [M, N)}
  \For{each instance $x_i$}
    \State sample $s$ captions $\widetilde{y}_1, \dots, \widetilde{y}_s$
    \For{each sample $\widetilde{y}_j$}
      \State calculate $\mathcal{L}^j(\theta)$ as in eq~\eqref{eq:surrogate_loss_sample}
    \EndFor
    \State calculate surrogate loss $\mathcal{L}(\theta)$ as in eq~\eqref{eq:surrogate_loss}
    \State update parameter $\theta$ by stochastic gradient descent
  \EndFor
\EndFor
\end{algorithmic}
\end{algorithm}

\subsection{Network Architecture and Training Details}
Our proposed objective and solution is compatible with any captioning model that follows the encoder-decoder architecture \cite{google}. 
The encoder depends on the input (image or video) and will be specified in the experiment section. 
The decoder is an RNN model of LSTM cell with hidden dimension set to $512$. 
We add one full connection layer after the encoder to reduce the dimension to $512$. 
In step $0$, the hidden state is initialized by the output of this full connection layer. 

We use CIDEr metric to calculate $\Delta(y, y_i)$ and we use BLEU3 + BLEU4 to calculate $d(y, y')$ in eq~\eqref{eq:surrogate_loss}. 
We set the number of samples $s$ to $5$. 
To reduce the variance introduced in the Monte Carlo sampling step when estimating the gradient in optimization, we follow the standard practice of using baseline. 
For the gradient of precision term, we set its baseline to the CIDEr score of greedily decoded caption from the model following work \cite{sc}. 
For the gradient of diversity term, we set it to $\frac{1}{s^2}\sum_{k=1}^s \sum_{j=1}^s d(\widetilde{y}_j, \widetilde{y}_k)$, the average of all the pairwise distances between sampled captions. 
We use ADAM optimizer in optimization. 

\section{Experiment}
\label{sec:expr}
In this section, we first introduce the experiment setup. 
Then we report the performance of the model trained by our proposed objective on standard evaluation metrics of precision side in the image captioning task and video captioning task respectively. 
Finally, we discuss the model behavior on both precision and recall sides. 

\subsection{Experiment Setup}
For the image captioning task, we use the MSCOCO dataset \cite{mscoco}, which is one of the largest image caption datasets that contains more than 120K images crawled from Flickr. Each image is annotated with 5 reference captions. We use the public split \cite{karpathy} for experiments. For the video captioning task, we use the TGIF dataset \cite{tgif}, which is one of the largest video caption datasets that contains 100K animated GIFs collected from Tumblr and $120K$ caption sentences. We use the official split \cite{tgif} for experiments.

For image, we use Resnet152 \cite{resnet} pretrained on ImageNet \cite{imagenet} and apply spatial mean pooling to get a $2048$-dim feature vector. 
For video, we also use Resnet152 \cite{resnet} for fair comparison to other works rather than use a stronger CNN such as I3D \cite{i3d}. 
We apply spatial-temporal mean pooling to get a $2048$-dim feature vector. 
For simplicity, we don't finetune the feature on the caption datasets. 
We tune the hyper-parameter $\alpha$ in eq~\eqref{eq:problem} among $.25$, $.5$ and $.75$ on the validation set and set it to $.75$. 
We find that $.75$ is a quite stable value to reach the best performance across different datasets. 

\subsection{Image Captioning}
We first study the contribution of our proposed objective by comparing it to training our model with the original sequence-level learning loss (SLL) and sequence-level learning with maximum entropy regularization (SLL-ME) \cite{entropy_rl}. 
The weight of the maximum-entropy regularization in SLL-ME is tuned among $10^{-1}$, $10^{-2}$, $10^{-3}$ and set to $10^{-2}$ for the best performance. 
Both the network architecture and input feature are the same across SLL, SLL-ME and SLL-SLE (ours). 
We use beam search in test stage with width of $5$. 
As shown in the middle block from table~\ref{tab:image_caption},  we can see that our model SLL-SLE improves over SLL and SLL-ME significantly on all metrics. 
The improvement of SLL-SLE over SLL-ME on all metrics (Meteor: $0.2$, CIDEr: $1.8$, SPICE: $0.2$) is much larger than the improvement of SLL-ME over SLL on all metrics (Meteor: $0.0$, CIDEr: $0.6$, SPICE: $0.1$). 
This shows that the typical maximum-entropy regularization doesn't help to solve the issue of original sequence-level objective in the captioning task. 
Our proposed sequence-level exploration is effective in guiding the model to explore more plausible captions in training and consequently SLL-SLE generates more accurate captions in test. 
In the last block of table~\ref{tab:image_caption}, we also include results of SLL, SLL-ME, SLL-SLE objectives when combined with attention architecture. 
Again the similar trend is observed: SLL-SLE improves over SLL and SLL-ME significantly. 

We also compare our proposed model to various state-of-the-art (SOTA) models with different network architectures trained by either word-level cross-entropy loss or sequence-level learning objective. 
For word-level XE loss, we compare to NIC model \cite{google}, Adaptive \cite{adaptive_attention}, Top-down attention \cite{bottom_up}. 
For sequence-level learning objective (SLL), we compare to self-critical learning (SCST:FC \& SCST:Att2in) \cite{self_critique} and Top-Down attention \cite{bottom_up}. 
As shown in table~\ref{tab:image_caption}, we see that the proposed objective leads to better performance on all metrics over all SOTA models. 

\begin{table}[!t]
\centering
\caption{Performance improvement on the image captioning: * means bottom-up region features are used with attention architecture}
\begin{tabular}{cccc}\toprule
Method & Meteor & CIDEr & Spice \\\cmidrule(lr){1-1}\cmidrule(lr){2-4}
NIC \cite{google} & $23.7$ & $85.5$ & NA\\
Adaptive \cite{adaptive_attention} & $26.6$ & $108.5$ & NA\\ 
SCST:FC \cite{self_critique} & $25.5$  & $106.3$ & NA\\
SCST:Att2in \cite{self_critique} & $26.3$ & $111.4$ & NA\\
Top-Down-XE \cite{bottom_up} & $26.1$ & $105.4$ & $19.2$ \\
Top-Down-SLL \cite{bottom_up} & $26.5$ & $111.1$ & $20.2$ \\
\midrule
SLL & $26.8$ & $115.0$ & $20.0$ \\
SLL-ME & $26.8$ & $115.6$ & $20.1$ \\
SLL-SLE (ours) & $\mathbf{27.0}$ & $\mathbf{117.2}$ & $\mathbf{20.3}$ \\\midrule
SLL* & $26.6$ & $117.2$ & $19.4$ \\
SLL-ME* & $26.7$ & $117.9$ & $19.5$ \\
SLL-SLE* (ours) & $\mathbf{27.0}$ & $\mathbf{119.6}$ & $\mathbf{19.9}$ \\\bottomrule
\end{tabular}
\label{tab:image_caption}
\end{table}

\subsection{Video Captioning}
Similarly, we first compare our proposed objective with original sequence-level learning loss (SLL) and sequence-level learning with maximum entropy regularization (SLL-ME). 
As we fix the hyper-parameter across datasets for our method (SLL-SLE), we also fix the hyper-parameter (weight before maximum-entropy regularization) in SLL-ME and set it to $10^{-2}$, same as that on MSCOCO dataset. 
We use beam search with width of $5$ in test stage. 
As shown in the last three rows from table~\ref{tab:video_caption}, we can see that our model, SLL-SLE, again improves over SLL and SLL-ME significantly on all metrics. 
Actually, SLL-ME performs worse than SLL on all metrics, which indicates that the maximum-entropy regularization is not stable across datasets and may even deteriorate the performance in some captioning task. 
Our model, SLL-SLE improves over SLL by $0.6$ on Meteor, $2.7$ on CIDEr and $0.6$ on SPICE with the same hyper-parameter setting as that on MSCOCO. 
This shows that the proposed sequence-level exploration term is stable and robust across datasets and are helpful to the model performance in general. 

\begin{table}
\centering
\caption{Performance improvement on the video captioning}
\begin{tabular}{cccc}\toprule
Method & METEOR & CIDEr & SPICE \\\cmidrule(lr){1-1}\cmidrule(lr){2-4}
Official\cite{tgif} & $16.7$ & $31.6$ & NA \\
Show-adapt\cite{show_adapt} & $16.2$ & $29.8$ & NA\\
\midrule
SLL & $17.8$ & $45.9$ & $15.9$ \\
SLL-ME & $18.2$ & $48.1$ & $16.0$ \\
SLL-SLE (ours) & $\mathbf{18.8}$ & $\mathbf{50.8}$ & $\mathbf{16.6}$ \\\bottomrule
\end{tabular}
\label{tab:video_caption}
\end{table}

We also compare our proposed model to various state-of-the-art (SOTA) models on the video captioning task. 
The TGIF dataset comes with an official baseline (Official) \cite{tgif} trained by word-level cross-entropy loss. 
Show-adapt \cite{show_adapt} leverages both TGIF and other datasets in training. 
By comparing our implementation of baseline model SLL to these models, we see that it performs better than them, which indicates that SLL is already a very strong baseline. 
This further suggests that the improvement over SLL is not trivial.

\begin{table}[!t]
\centering
\caption{Comparison of models trained by XE, SLL, SLL-ME, our SLL-SLE on both precision and diversity sides (MSCOCO dataset): (rs) denotes random sampling decoding and (bs) denotes beam search decoding}
\label{tab:SC_set}
\setlength{\tabcolsep}{3pt}
\begin{tabular}{ccccc}\toprule
\multirow{2}{*}{Method} & precision & \multicolumn{3}{c}{recall}\\
& CIDEr & Div1 ($\uparrow$) & Div2 ($\uparrow$) & mBleu4 ($\downarrow$)  \\\cmidrule(lr){1-1}\cmidrule(lr){2-2}\cmidrule(lr){3-5}
XE (rs) & $74.2$ & $0.57$ & $0.78$ & $0.06$ \\
SLL (rs) & $114.6$ & $0.25$ & $0.32$ & $0.81$ \\
SLL-ME (rs) & $115.1$ & $0.25$ & $0.33$ & $0.80$ \\
SLL-SLE (rs) & $115.9$ & $0.29$ & $0.40$ & $0.68$ \\\midrule
XE (bs) & $102.5$ & $0.27$ & $0.35$ & $0.80$ \\ 
SLL (bs) & $115.0$ & $0.26$ & $0.35$ & $0.78$\\
SLL-ME (bs) & $115.6$ & $0.26$ & $0.34$ & $0.79$ \\
SLL-SLE (bs) & $\mathbf{117.2}$ & $\mathbf{0.27}$ & $\mathbf{0.36}$ & $\mathbf{0.76}$ \\\midrule
VAE\cite{vae} (bs) & $100.0$ & NA & NA & NA \\
GAN\cite{caption_gan} (rs) & NA & $0.41$ & $0.55$ & $0.51$\\
GAN\cite{caption_gan} (bs) & NA & $0.34$ & $0.44$ & $0.70$ \\\bottomrule
\end{tabular}
\end{table}

\subsection{Discussion of Model Behavior on Precision and Recall}
We study the model behavior on precision and recall sides for these objectives: cross-entropy (XE), sequence-level learning (SLL), sequence-level learning with maximum-entropy (SLL-ME), our SLL-SLE. 
On the precision side, we use CIDEr metric as it is shown to have good correlation with human judgement. 
On the recall side, we use diversity metrics Div1, Div2, mBleu\cite{caption_gan} as proxy measurements. 
To calculate the diversity metrics, we adopt two decoding straties as \cite{caption_gan}. 
The first decoding strategy is to sample $5$ captions from the model for each image (rs). 
The second decoding strategy is to beam search top $5$ captions from the model for each image (bs). 
The reported CIDEr is the average of CIDEr scores of the $5$ sampled captions. 
As shown in table~\ref{tab:SC_set}, compared to SLL and SLL-ME,  the proposed objective, SLL-SLE, performs not only better on the precision side and but also better on the recall side under both random sampling and beam search decoding strategies. 
Compared to XE, SLL-SLE improves on both precision and recall aspects under beam search decoding strategies. 
We also list VAE and GAN's performance on precision and recall aspects for reference. 

\begin{figure}[!t]
\centering
\begin{subfigure}{.85\linewidth}
\includegraphics[width=\linewidth]{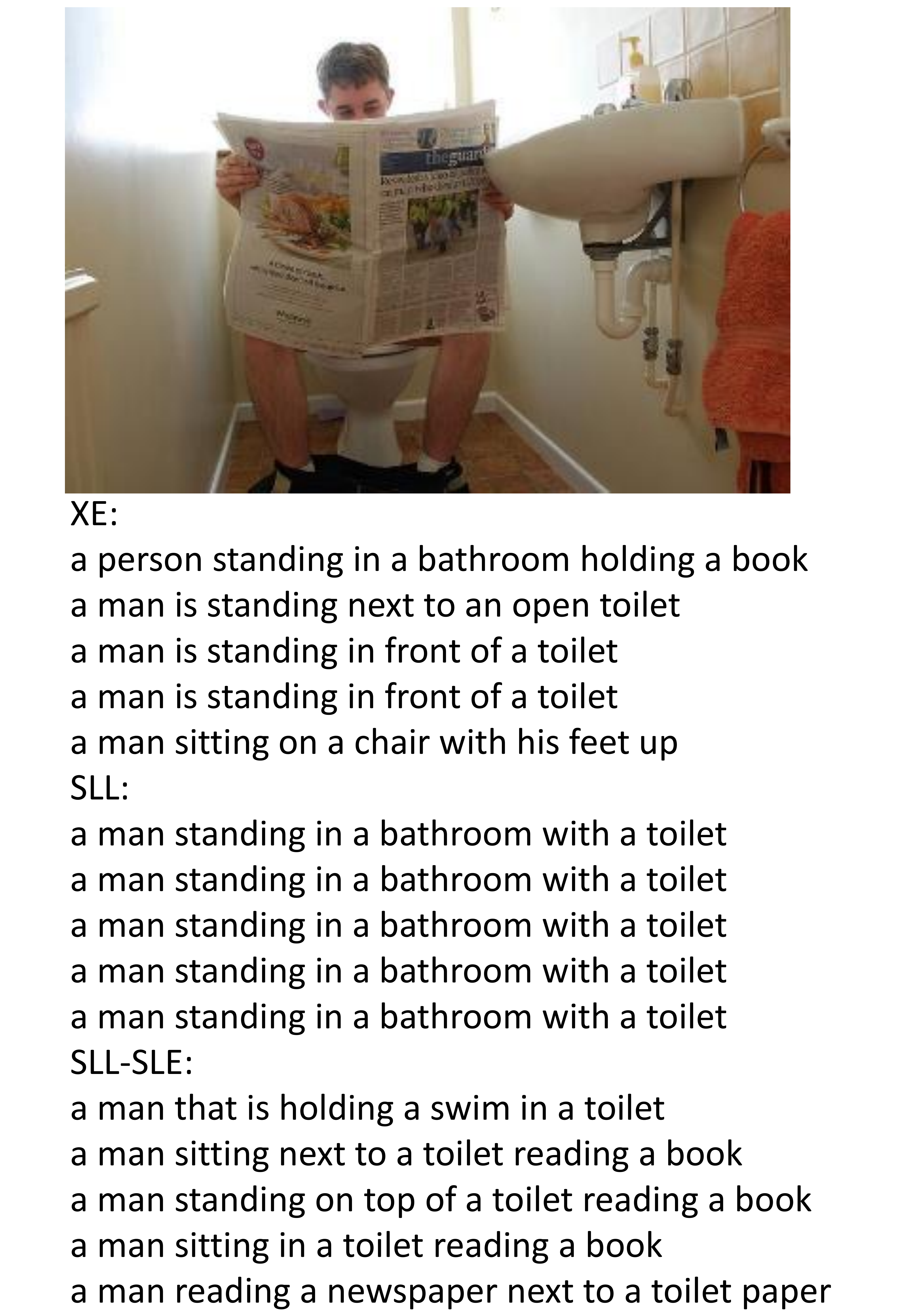}
\end{subfigure}
\caption{Case study of model behavior on precision and recall by sampling strategy in decoding}
\label{fig:diversity}
\vspace{-5pt}
\end{figure}

Figure~\ref{fig:diversity} shows that the proposed objective can generate diverse and high quality captions with sampling strategy. The quality of captions generated by the XE model is not good. The SLL model with sampling strategy has limited diversity and keeps generating almost the same caption with sampling strategy. 




\section{Conclusion}
\label{sec:conclusion}
In this work, we show the limitation of current sequence-level learning objective in captioning tasks from both theoretical and empirical aspects. 
From the theoretical aspect, this objective is equivalent to maximizing the generalized precision of the predicted caption set, which ignores the recall side. 
From the empirical aspect, models trained by this objective receive low score on proxy measurements of recall. 
To overcome the above limitations, we propose adding a sequence-level exploration term to maximize the diversity, a proxy measurement of recall, on generated captions. 
It encourages the model to explore more captions that are different in syntax but are semantically coherent with the groundtruth in training. 
Extensive experiments on both image and video captioning tasks show that the proposed objective leads to a win-win solution that consistently performs better on both precision and recall.

\section{Acknowledgement}
We would like to express our great appreciation to Shiwan Zhao for insightful discussions and valuable suggestions. 
This work was partially supported by National Natural Science Foundation of China (No. 61772535) and Beijing Natural Science Foundation (No. 4192028).

{\small
\bibliographystyle{ieee_fullname}
\bibliography{egbib}
}

\end{document}